\documentclass[sigconf]{acmart}
\AtBeginDocument{%
  }

\copyrightyear{2026}
\acmYear{2026}
\setcopyright{cc}
\setcctype{by}
\acmConference[CIKM '26]{Proceedings of the 35th ACM International Conference on Information and Knowledge Management}{November 07--11, 2026}{Rome, Italy}
\acmBooktitle{Proceedings of the 35th ACM International Conference on Information and Knowledge Management (CIKM '26), November 07--11, 2026, Rome, Italy}
\acmDOI{10.1145/3799682.3840265}
\acmISBN{979-8-4007-2539-5/2026/11}

\begin{document}

\title{ShanLiangRen: A Nutrition Agent for Personalized Daily Meal Planning}

\author{Miao Xie}
\email{0520shui@163.com}
\orcid{0000-0002-9569-509X}
\correspondingauthor
\affiliation{%
  \institution{College of Information and Electrical Engineering, China Agricultural University}
  \city{Beijing}
  \country{China}
}

\author{Xiao Zhang}
\email{B20243080802@cau.edu.cn}
\affiliation{%
  \institution{College of Information and Electrical Engineering, China Agricultural University}
  \city{Beijing}
  \country{China}
}

\author{Yuan Wang}
\email{wy929\_cs@163.com}
\affiliation{%
  \institution{College of Information and Electrical Engineering, China Agricultural University}
  \city{Beijing}
  \country{China}
}

\author{Ruixin Zhu}
\email{zhuruixin07@126.com}
\correspondingauthor
\affiliation{%
  \institution{Key Laboratory of Precision Nutrition and Food Quality, China Agricultural University}
  \city{Beijing}
  \country{China}
}

\author{Chunli Lv}
\email{lvcl@cau.edu.cn}
\correspondingauthor
\affiliation{%
  \institution{College of Information and Electrical Engineering, China Agricultural University}
  \city{Beijing}
  \country{China}
}

\renewcommand{\shortauthors}{Miao Xie, Xiao Zhang, Yuan Wang, Ruixin Zhu, and Chunli Lv}

\begin{abstract}
Dietary nutrition planning plays an important role in chronic disease management and maintaining a healthy body. In applications, it must simultaneously satisfy personalized constraints and reasonable multidimensional nutritional goals. These two aspects often conflict, and user constraints evolve with feedback, resulting in a substantial gap between generic guidelines and executable plans. 
    To bridge this gap, we first propose the personalized \textsc{fully quantified multiobjective dietary planning problem (MDP)}. To tackle \textsc{MDP}, we develop a nutrition agent, \textsc{ShanLiangRen}.
    The system first transforms dietary specifications, nutrient data, user attributes and natural language requirements into an individualized constrained planning instance. It then employs an exact retrieval-augmented generation method to shrink the feasible candidate set from a large scale ingredient and recipe space. Finally, it adopts a refinement guided by Pareto principles, where an LLM iteratively revises candidate plans under deterministic nutrition computation and feedback from constraint verification. The system outputs fully quantified meal plans with explicit ingredients and portion sizes, together with reports on nutrition compliance that show constraint satisfaction and nutrient interval attainment.
    We have released the system online as a WeChat Program, \textsc{ShanLiangRen}. A demo video is available at \url{https://www.youtube.com/watch?v=652OtY5VlGA}.
\end{abstract}




\maketitle

\section{Introduction}
Unhealthy diets have become an important driving factor of the global burden of chronic diseases: a systematic analysis covering 195 countries estimated that in 2017 about 11 million deaths were associated with dietary risks and caused about 255 million Disability-Adjusted Life Years (DALYs)~\cite{20191959}. Meanwhile, an increasing number of studies indicate that more scientific and individualized dietary interventions can affect metabolic health and disease risk\cite{b14,xie2026precisionnutrition}.
Existing dietary-planning studies follow two main directions. Health-oriented studies target chronic-disease risk factors, such as obesity and abnormal glucose levels, and develop intervention schemes or clinical nutrition management workflows centered on nutritional goals and food combinations~\cite{PMID:36521448,MEHTA2024100186,b11}. Computational studies model menus or recipes as combinatorial optimization or recommendation problems under nutritional and preference constraints, enabling the automatic generation of executable diet plans~\cite{10.1145/3394486.3403223,DBLP:conf/ijcai/0001ZG0MC22,DBLP:conf/ijcai/TianZGMMC22,9942937,SAHIN2024124213,10.1145/3712709}.
However, two limitations remain: (1) personalized constraints evolve with users' physiological states and often conflict with reasonable nutritional goals, making guideline-based plans difficult to follow and execute at the individual level; and (2) natural-language requirements lack a unified representation and must be reliably normalized into executable constraints and goals, while open-ended dialogue often drifts toward general nutrition Q\&A with limited planning actionability~\cite{b12,10.1145/3712709}.
Recent studies have explored integrating LLMs with structured nutrition knowledge or multiobjective optimization to support personalized and explainable dietary recommendations~\cite{gao2025healthgenie,zhang2025mopi}.

Accordingly, we formulate the \textsc{fully quantified multiobjective dietary planning problem (MDP)}. Given dietary specifications and user information, MDP aims to dynamically generate a plan suited to the user's physical condition, quantify its ingredients and portions, and address multidimensional targets beyond calories and macronutrients.
Solving \textsc{MDP} involves three coupled challenges. First, the system must formulate an individualized planning instance by converting general dietary specifications into nutrient target intervals based on the user's physical attributes, while incorporating stable restrictions and daily preferences as requirements or priorities. Second, it must construct feasible meal candidates from structured food and recipe tables with deterministic nutrient composition data, ensuring reliable evidence for ingredients, portions, and nutrient values. Third, it must refine feasible candidates across multiple nutritional objectives and personalized preferences, using deterministic nutrient computation to verify compliance with the target intervals.

To address \textsc{MDP}, we develop \textsc{ShanLiangRen}, an interactive system for personalized daily dietary planning. It focuses on constructing feasible candidates from a large ingredient space and producing executable solutions under multiple objectives through two key designs. First, an exact RAG mechanism based on subgraph matching reduces the feasible search space and generates candidate plans. It represents dietary specifications and user constraints as retrieval filters, abstracts meal structure, category coverage, and combination principles into a query subgraph, and retrieves combinations satisfying user requirements through exact structural matching over the knowledge graph. Second, a Pareto guided multiobjective refinement strategy balances constraint satisfaction and nutritional goals. Through a feedback loop of deterministic nutrient computation and constraint verification, it evaluates and optimizes the candidates, improving nutritional objectives while satisfying personalized constraints, and outputs the recommended plan.

This demo makes three contributions: 
(1) we define the \textsc{personalized fully quantified multiobjective dietary planning problem (MDP)}; 
(2) we design an exact retrieval and Pareto guided refinement pipeline  with deterministic nutrient verification; 
and (3) we develop a system for personalized daily meal planning.

\section{Problem Definition}
We represent a dietary planning task as an instance $I=(S,T,u)$. 
The planning instance is grounded in external nutritional knowledge rather than generated nutrition claims. 
Specifically, $S=(G,P)$ denotes the dietary specifications, where $G$ is derived from the Chinese Dietary Guidelines (2022) and $P$ follows the Chinese Food Guide Pagoda (2022)~\cite{cns2022foodguide}. 
$T$ is the ingredient nutrient table built from the China Food Composition Tables~\cite{chinafood2018v1,chinafood2019v2}, and $u=(u_{\text{attr}},u_{\text{req}})$ is the user information. 
Here, $u_{\text{attr}}$ describes the user's basic physical signs and health attributes, and $u_{\text{req}}$ describes the user's dynamic requirements. 
Therefore, the LLM is used to normalize user requirements and refine candidate plans, while nutrient values, target intervals, and constraint verification are grounded in external specifications and structured data.
The system outputs a fully quantified daily plan $x=\{x_m \mid m\in M\}$, where $M$ is the set of meals, and each $x_m$ gives the dishes/ingredients and their portion sizes. 
A feasible solution must satisfy two types of hard constraints: $x \models C_u(u_{\text{req}})$ and $x \models C_{\text{qual}}(S)$, corresponding respectively to the personalized requirements that the user must satisfy and the qualitative healthy structural principles that the specifications must follow. 
Based on $T$, we compute the daily intake nutrition vector $n(x)=\mathrm{Compute}(x;T)$, and derive the individualized nutrition recommendation set $R(u_{\text{attr}},S)=\psi(u_{\text{attr}},S)$ from $u_{\text{attr}}$ and $S$, based on which we construct the nutrition objective set $J_{\text{nut}}(S,u_{\text{attr}})$.
Here $C_u(u_{\text{req}})$ mainly covers individual hard restrictions such as allergies, taboos, meal structure requirements, and explicit daily preferences, while $C_{\text{qual}}(S)$ captures dietary structure principles derived from guidelines such as category coverage and meal balance.

For each $j\in J_{\text{nut}}(S,u_{\text{attr}})$, $R(u_{\text{attr}},S)$ provides its reasonable intake interval $[L_j,U_j]$. To characterize the deviation degree of plan $x$ on objective $j$ relative to this interval, we define the interval deviation:
\begin{equation}
d_j(x)=\frac{\max\{0,\,L_j-n_j(x),\,n_j(x)-U_j\}}{\max\{\varepsilon,\,U_j-L_j\}},
\end{equation}
where $n_j(x)$ denotes the intake value of plan $x$ on nutrition objective $j$, and $\varepsilon$ is a very small constant.
Under the premise of satisfying the hard constraints, the goal is to make all nutrition objectives fall into their reasonable intervals as much as possible, and minimize the overall deviation:
\begin{equation}
\begin{aligned}
\min\ \ L(x) &= \sum_{j\in J_{\text{nut}}(S,u_{\text{attr}})} w_j\cdot d_j(x) \\
\text{s.t.}\ \ & x \models C_u(u_{\text{req}}),\quad x \models C_{\text{qual}}(S).
\end{aligned}
\end{equation}
where $w_j$ denotes the importance weight of nutrition objectives.

\section{System Design}

\begin{figure}[!t]
\includegraphics[width=0.99\columnwidth]{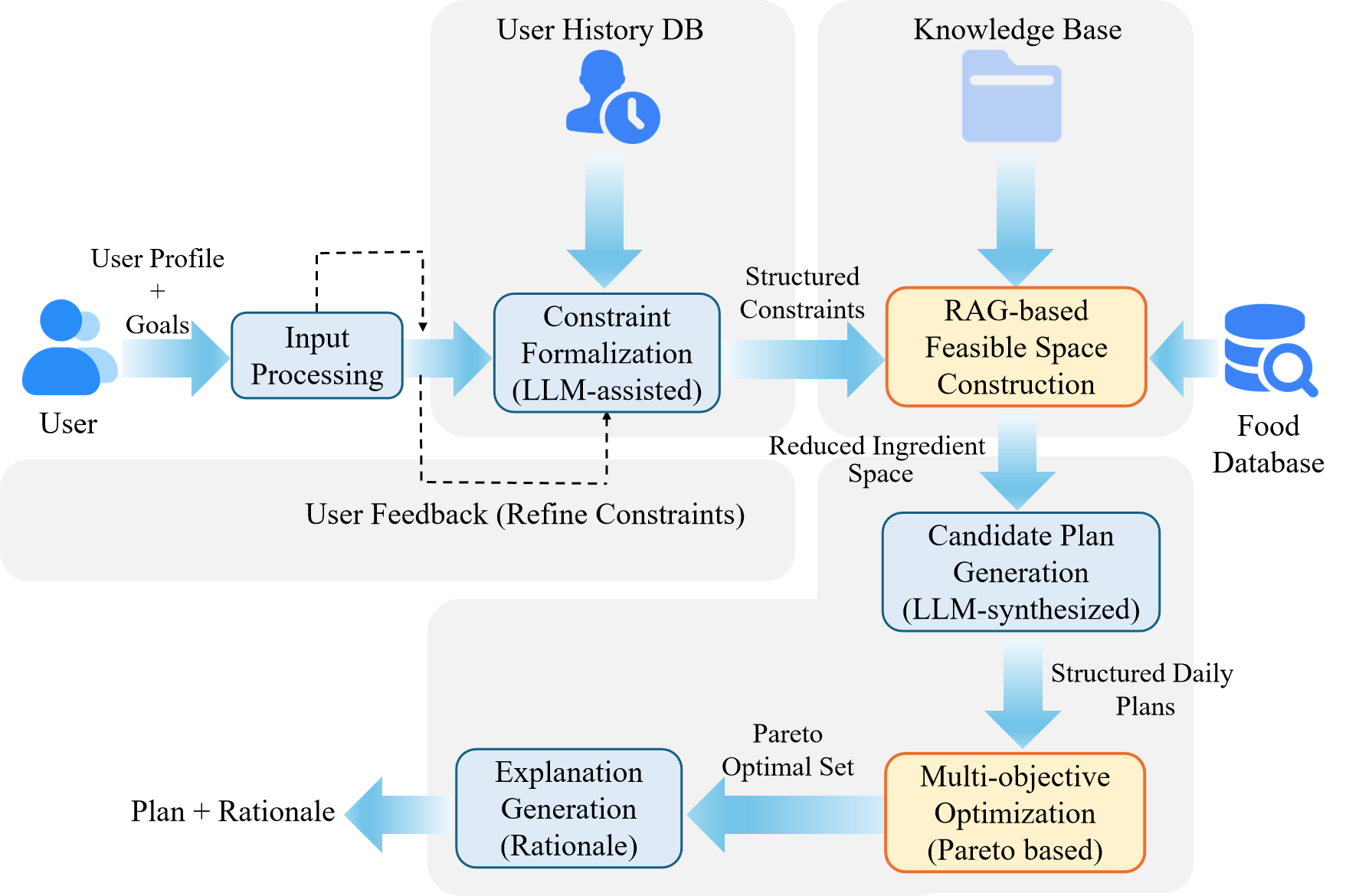}
\caption{Overall architecture of the ShanLiangRen system.}
\label{fig1}
\end{figure}

\textbf{Overview.}
Given an \textsc{MDP} instance, \textsc{ShanLiangRen} follows a five module online pipeline (Fig.\ref{fig1}): (1) individual constraint and objective formulation, (2) exact retrieval for feasible space reduction, (3) quantified candidate generation, (4) deterministic nutrition evaluation, and (5) Pareto guided refinement with explanation.
\textbf{Implementation.} ShanLiangRen employs GPT-5.4 as its LLM backbone. The LLM is responsible for requirement normalization, candidate proposal, and targeted plan repair.

\textbf{\textit{(1) Dynamic Constraint Extraction and Task Structuring.}}
The system first receives user information $u=(u_{\text{attr}},u_{\text{req}})$. The LLM performs structured extraction on $u_{\text{req}}$ to form executable user constraints $C_u(u_{\text{req}})$, covering hard requirements such as taboos/allergies, taste preferences, and meal/course structure (e.g., number of meals, number of dishes per meal, no repetition)\cite{DBLP:conf/iclr/YaoZYDSN023,DBLP:conf/nips/SchickDDRLHZCS23}. Meanwhile, the system constructs the individualized nutrition objective set $J_{\text{nut}}(S,u_{\text{attr}})$ based on $u_{\text{attr}}$ and the specifications $S$, and generates a recommended intake interval $[L_j,U_j]$. This step uniformly maps user inputs to the formal objects in the problem definition, providing a consistent interface for retrieval reduction and plan generation. In our demo, \textsc{ShanLiangRen} adopts a controlled, plan based interaction design. Stable user attributes and stable restrictions are collected through structured fields, while daily preferences and revision requests can be expressed in natural language with respect to the current meal plan. This design preserves personalized expressiveness while preventing the interaction from drifting into open-ended nutrition consultation.

\textbf{\textit{(2) Exact RAG for Feasible Space Reduction.}}
To avoid combinatorial search over the full library, the system adopts exact RAG based on subgraph matching to reduce the candidate space to $R_u$. \cite{DBLP:conf/nips/LewisPPPKGKLYR020,DBLP:journals/corr/abs-2002-08909,DBLP:conf/sigir/KhattabZ20,xie2026structure,zhang2026mcrag} Specifically: (1) \textbf{exact filtering of hard constraints}: it first removes items that violate $C_u(u_{\text{req}})$ in the ingredient/recipe library (e.g., forbidden ingredients, infeasible meal structures), and combines the qualitative specification constraints $C_{\text{qual}}(S)$ for structural elimination (e.g., basic feasibility of category coverage/combination templates). 
For example, a requirement such as avoiding a specific ingredient, including a vegetable category in lunch, and preventing repeated staple foods can be converted into exclusion predicates at the ingredient level, category coverage constraints, and meal structure relations. Exact retrieval means that candidates violating these hard predicates or structural relations are excluded before LLM generation, so the LLM only operates on a user relevant feasible subspace.
(2) \textbf{subgraph matching retrieval}: it abstracts ``meal structure/category coverage/combination principles'' as a query graph, and performs exact matching retrieval under structural constraints on the filtered set to obtain the candidate subspace most relevant to the current user context, yielding $R_u \subseteq D$. After this reduction, subsequent LLM generation and refinement are performed only within $R_u$. Unlike conventional RAG that primarily retrieves semantically relevant textual contexts, our exact RAG retrieves structured meal candidates that explicitly satisfy query graph relations. If no feasible candidate satisfies all hard constraints, ShanLiangRen does not silently relax safety-critical requirements. Instead, the system reports the infeasible constraint combination and requests revision of lower-priority daily preferences before replanning.

\textbf{\textit{(3) Candidate Generation.}}
Given $R_u$, the LLM generates a diverse set of candidate daily plans $\{x^{(1)},\ldots,x^{(K)}\}$~\cite{DBLP:conf/iclr/0002WSLCNCZ23}. Each candidate explicitly specifies meal partitioning and ingredient/dish portion sizes, ensuring ``fully quantified executability''. 
Since the generation is constrained by $R_u$, candidate plans inherit the retrieved feasibility conditions and are easier to verify and repair than unconstrained LLM outputs.
The generation stage enforces structural hard requirements and promotes diversity for trade-off selection.
\textbf{Deterministic Evaluation.}
For each candidate $x$, the evaluator computes the daily intake vector $n(x)=\mathrm{Compute}(x;T)$ based on the element table $T$, and checks whether hard constraints are satisfied ($x \models C_u(u_{\text{req}})$ and $x \models C_{\text{qual}}(S)$). Then, according to the recommended interval $[L_j,U_j]$ of each objective, the system computes the deviation $d_j(x)$ and aggregates it to obtain the overall deviation $L(x)$.
The resulting deficit, excess, and interval attainment status are used both for user explanation and for the next refinement round. The LLM serves as a semantic proposal operator that combines retrieved ingredients according to natural-language preferences and produces targeted modifications while preserving unaffected constraints.

\textbf{\textit{(4) Pareto Guided Refinement with LLM.}}
Unlike traditional Pareto optimization, we agentize the candidate generation and repair operators: the LLM performs iterative ``minimal modification'' driven by deterministic evaluation feedback, rather than using handcrafted search operators~\cite{DBLP:conf/nips/MengLPC22}. The system maintains an evaluated candidate pool and performs dominance filtering based on the deviation vector $d(x)=\{d_j(x)\}$ (the smaller the better): it keeps candidate solutions that are no worse on all objective deviations and strictly better on at least one objective deviation, retains candidates reflecting different trade-offs, and removes dominated solutions that are not better on any considered dimension. Then, the system feeds the interval deviation feedback produced by the evaluator into the LLM, and the LLM performs ``minimal-modification'' repairs, entering the next round until reaching an iteration limit or the improvement becomes stable.

\textbf{\textit{(5) Recommendation and Explanation.}}
After the candidate pool converges, the system outputs: (1) the recommended plan $x^\ast$, selected from the non-dominated set according to the user's preference priorities; (2) an explainable report, showing the satisfaction of key hard constraints, nutrition computation results $n(x)$, interval attainment for each objective (whether it falls into $[L_j,U_j]$) and the corresponding deviation $d_j(x)$/deficit or excess.

\begin{figure}[!h]
\includegraphics[width=0.99\columnwidth]{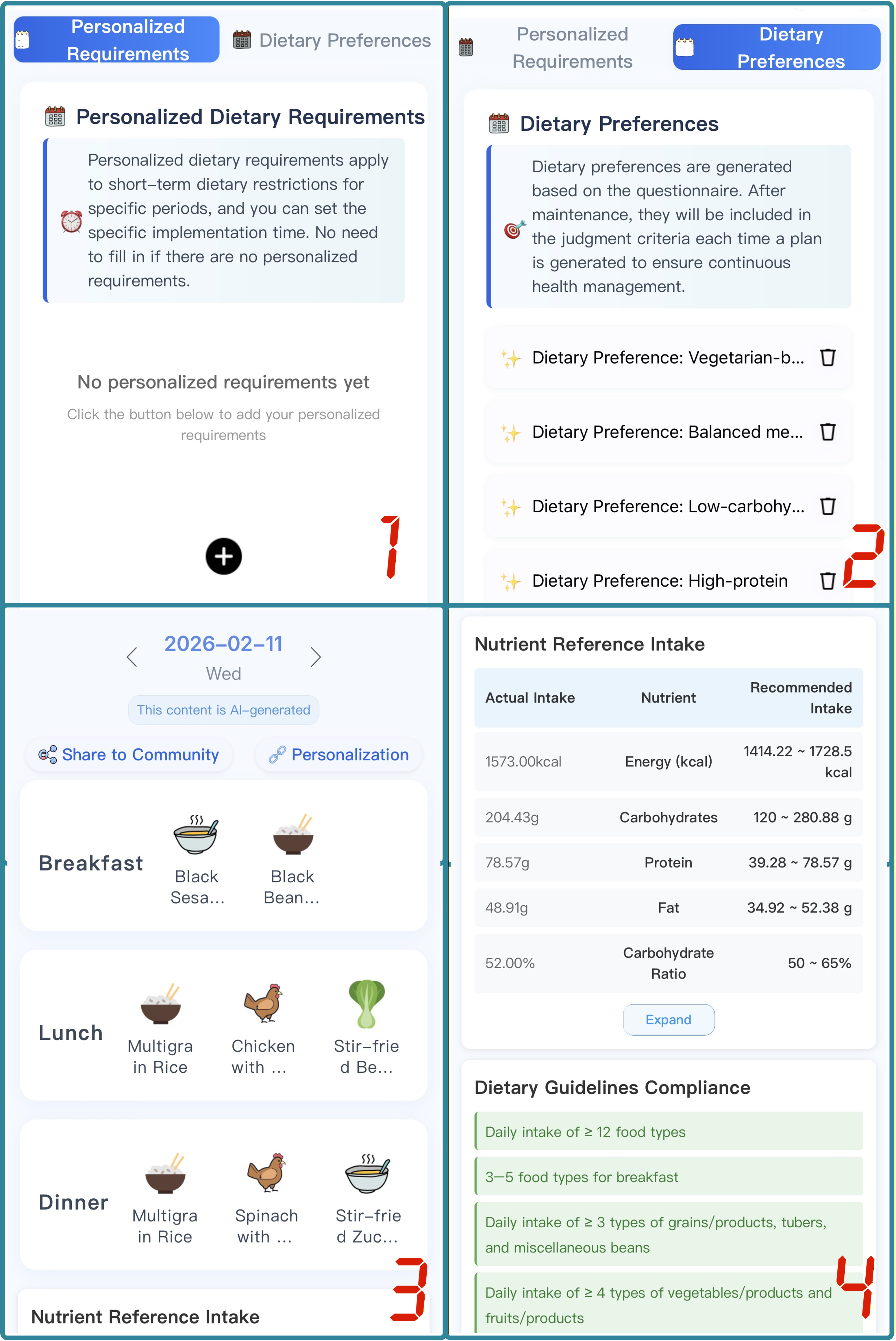}
\caption{Interface of the ShanLiangRen system.}
\label{fig2}
\end{figure}
\textbf{Backend Knowledge Base.}
\textsc{ShanLiangRen} maintains a structured ingredient--recipe knowledge base in the backend, and uses the ingredient nutrition element table as the computation basis of the deterministic evaluator\cite{DBLP:conf/ijcai/TianZGMMC22}. The knowledge base contains structured information including recipe entities, ingredient entities, ingredient category hierarchies, and macro and micro nutrient attributes, enabling the system to abstract requirements such as “meal-type structure/category coverage/combination principles” into query subgraphs and to perform exact matching retrieval within the candidate subspace. \textbf{Data Scale:} the current backend includes $1,550,150$ recipes and $64,127$ ingredients; each ingredient is annotated with 64 nutrient attribute dimensions, and a category hierarchy is provided.
The backend knowledge base supports two different functions: structured retrieval and deterministic verification. Recipe and ingredient entities are used to construct candidate meal combinations under restriction and category coverage constraints, while the nutrient table is used to compute the actual intake values of generated plans.

\section{System in Action}

As shown in Fig.~\ref{fig2}, \textsc{ShanLiangRen} has been deployed as an online WeChat Mini Program, where users can access the system, submit dietary requirements, and receive plan based dynamic updates ~\cite{DBLP:conf/uist/ParkOCMLB23,DBLP:conf/iclr/ChenSZ0YCYLHQQC24}: it stores stable restrictions as stable constraints, and incrementally re-plans when users revise daily needs via structured options and comments on the current plan.
On the output side, the system generates a fully quantified daily plan with three meals, refined to dishes/ingredients and portion sizes, and performs deterministic nutrition computation based on the ingredient element table, providing interval comparisons for indicators such as energy and macronutrients as well as attainment verification of requirements from dietary specifications, and finally presents an executable personalized daily plan in the form of ``recommended plan + explainable verification report''. In the demo, the user first fills in basic physical signs and stable preferences (Fig.\ref{fig2}.2), and inputs daily requirements (Fig.\ref{fig2}.1); the system immediately parses them into structured hard constraints and generates a fully quantified three-meal daily plan (Fig.\ref{fig2}.3). Then, the interface shows nutrition computation results and specification attainment verification (Fig.\ref{fig2}.4), and the user can choose among the recommended plan and several trade-off alternative plans. If the user temporarily adjusts requirements (e.g., taboos or meal structure changes), the system triggers fast incremental replanning and synchronously updates the explainable verification report.

\section{Evaluation}

We evaluate \textsc{ShanLiangRen} on $300$ distinct planning instances covering common personalized requirements, including ingredient taboos, dietary preferences, meal-structure constraints, and nutrition goals. We compare \textsc{ShanLiangRen} with direct LLM generation and constrained prompting, where all methods receive the same user profiles and daily requirements.
All generated plans are manually verified by a Ph.D. in nutrition based on the China Food Composition Tables~\cite{chinafood2018v1,chinafood2019v2}, and the verification results are used to compute five metrics. PCS is the percentage of satisfied individual hard constraints. EPC is the percentage of plans with complete dish, ingredient, and portion information. NIA is the percentage of checked nutrient objectives whose verified values fall within the recommended intervals. Plans lacking sufficient portion information are marked as incomplete, and their corresponding nutrient objectives are treated as unverifiable. AND is the average normalized nutrient deviation $L(x)$ in Eq.~(2), where lower is better. Latency is the average end-to-end generation time in seconds. 
\begin{table}[!h]
\centering
\caption{Overall comparison on 300 user planning cases.}
\label{tab:evaluation}
\begin{tabular}{lccccc}
\hline
Method & PCS & NIA & EPC & AND & Lat. \\
       & (\%)$\uparrow$ & (\%)$\uparrow$ & (\%)$\uparrow$ & (norm.)$\downarrow$ & (s)$\downarrow$ \\
\hline
GPT-5.4 & 87.5 & 60.4 & 85.9 & 0.17 & 46 \\
Gemini-3 & 75.0 & 55.2 & 80.2 & 0.25 & 19 \\
Doubao-seed 2.0 & 87.5 & 61.1 & 79.9 & 0.18 & 27 \\
\textbf{ShanLiangRen} & \textbf{100.0} & \textbf{78.9} & \textbf{100.0} & \textbf{0.09} & \textbf{61} \\
\hline
\end{tabular}
\end{table}

As shown in table~\ref{tab:evaluation},
\textsc{ShanLiangRen} achieves $100.0\%$ PCS and EPC, with the highest NIA of $78.9\%$ and the lowest AND of $0.09$, showing full individual constraint satisfaction, complete executable portion information, and lower overall nutrient deviation under full nutrition verification. GPT-$5.4$ and Gemini-$3$ provide partially complete portions but show lower PCS and NIA, while Doubao-seed $2.0$ performs competitively on NIA but has weaker portion completeness.Although \textsc{ShanLiangRen} incurs a moderate latency overhead, the additional time is acceptable for daily meal planning.

\section{Conclusion}
We presented \textsc{ShanLiangRen}, an interactive nutrition agent for personalized daily meal planning. We first defined the \textsc{personalized fully quantified multiobjective dietary planning problem (MDP)}, then designed an exact retrieval and refinement pipeline guided by Pareto principles with deterministic nutrient verification, and finally developed an interactive system that outputs executable meal plans with portion sizes and compliance reports. The evaluation shows that \textsc{ShanLiangRen} improves constraint satisfaction, nutrient attainment, executability, and verifiability over direct LLM generation, at the cost of acceptable additional latency.

\begin{acks}
This work was supported by the China Agricultural University ``Young Researcher'' Start-up Fund No.~QNYJY2024144 and the Visiting Scholar Program of the China Scholarship Council (CSC) No.~202506350123.
\end{acks}

\bibliographystyle{ACM-Reference-Format}
\bibliography{CIKM26}










\end{document}